\documentclass[10pt,twocolumn,letterpaper]{article}

\usepackage{3dv}
\usepackage{times}
\usepackage{epsfig}
\usepackage{graphicx}
\usepackage{amsmath}
\usepackage{amssymb}
\usepackage{fancyhdr}

\usepackage[pagebackref=true,breaklinks=true,letterpaper=true,colorlinks,bookmarks=false]{hyperref}

\threedvfinalcopy 

\def\threedvPaperID{157} 

\ifthreedvfinal\fi
\begin{document}

\title{Creating Realistic Ground Truth Data for the Evaluation of Calibration Methods for Plenoptic and Conventional Cameras}

\author{Tim Michels, Arne Petersen and Reinhard Koch\\
	Department of Computer Science, Kiel University, Germany\\
	{\tt\small \{tmi,arne.petersen,rk\}@informatik.uni-kiel.de}
}

\maketitle
\thispagestyle{fancy}
\renewcommand{\headrulewidth}{0pt}
\fancyfoot[C]{\footnotesize{$\copyright$ 2019 IEEE. Personal use of this material is permitted.
		Permission from IEEE must be obtained for all other uses, in any current or future
		media, including reprinting/republishing this material for advertising or promotional
		purposes, creating new collective works, for resale or redistribution to servers or
		lists, or reuse of any copyrighted component of this work in other works.
		DOI: \href{https://doi.org/10.1109/3DV.2019.00055}{10.1109/3DV.2019.00055}}}

\begin{abstract}
   Camera calibration methods usually consist of capturing images of known calibration patterns and using the detected correspondences to optimize the parameters of the assumed camera model. A meaningful evaluation of these methods relies on the availability of realistic synthetic data. In previous works concerned with conventional cameras the synthetic data was mainly created by rendering perfect images with a pinhole camera and subsequently adding distortions and aberrations to the renderings and correspondences according to the assumed camera model. This method can bias the evaluation since not every camera perfectly complies with an assumed model. Furthermore, in the field of plenoptic camera calibration there is no synthetic ground truth data available at all. We address these problems by proposing a method based on backward ray tracing to create realistic ground truth data that can be used for an unbiased evaluation of calibration methods for both types of cameras.
\end{abstract}

\section{Introduction}
The most commonly used camera calibration procedure consists of three steps: i) capturing images of calibration patterns, ii) detection of the patterns, \ie points of interest in the images belonging to the calibration pattern, and iii) using the correspondences to optimize the parameters of the assumed mathematical camera model. In order to evaluate single parts of this pipeline, synthetic calibration pattern renderings with known correspondences can be beneficial in two ways. Firstly, the quality of the pattern detection method can be assessed by comparing the detector results on the renderings to the ground truth positions, and secondly, the optimization as well as the camera model can be evaluated using the ground truth correspondences without depending on a possibly biased pattern detector. However, the validity of such an evaluation depends on the quality of the ground truth data, \ie its ability to reflect real data.\\
For conventional cameras the synthetic image generation is usually done by first rendering the calibration pattern from the perspective of a simple pinhole camera model so that the correspondences are easy to calculate. Afterwards the images and correspondences are then distorted according to the assumed camera model (see \eg \cite{zhang2000flexible}\cite{heikkila2000geometric}\cite{lucchese2002using}). This procedure poses the problem, that the generated data is not reflecting a real camera, but a virtual camera perfectly complying with the assumed distortion model. Accordingly, in a comparative evaluation of different calibration algorithms those methods assuming the exact same camera model have an advantage. Another problem is posed by the modeling of de-focus and image degradation effects like vignetting. In previous works these are either not modeled at all or simulated by adding Gaussian noise and blur to the perfectly distorted images. In neither of these cases the resulting images are directly comparable to real data resulting from a significantly more complex image formation process.\\
\begin{figure}[!t]
	\centering
	\includegraphics[width=.47\textwidth]{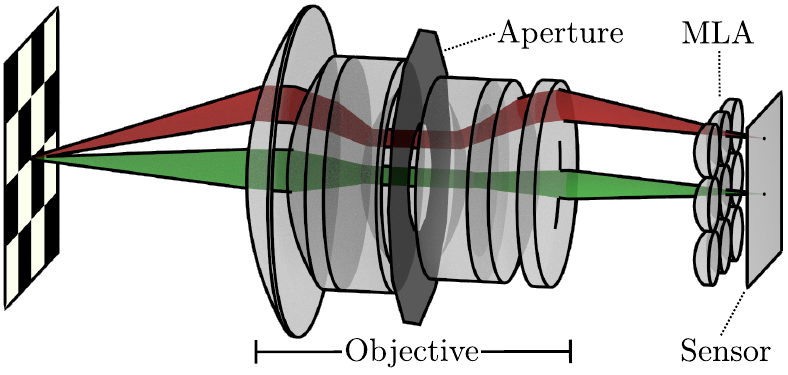}
	\caption{Schematics of a plenoptic camera as introduced by Adelson and Wang \cite{adelsonwang1992plenopticcam} and Lumsdaine and Georgiev \cite{lumsdaine2009focusedplenoptic} based on the ideas of Lippmann \cite{lippmann1908integralphoto}. The red and green cones indicate the areas visible from the corresponding pixels.}
	\label{fig:plenoptic}
\end{figure}\noindent
In the case of plenoptic cameras this imaging process is even more complicated since an additional microlens array (MLA) is placed between the main lens and the image sensor (compare \autoref{fig:plenoptic}). This renders the standard pipeline of creating perfect images and adding distortions and degradation afterwards infeasible since distortions of the main lens affect the position and angle of a ray hitting the MLA and therefore have to be applied before the light rays enter the microlenses.\\
We propose a pipeline to generate realistic renderings of calibration patterns with ground truth correspondences, \ie the positions of the calibration patterns' points of interest in the form of 2D pixel coordinates as well as 3D world coordinates. The key idea in generating these ground truth positions is to use ray tracing not only to render a realistic image $I$ of a calibration pattern, but also to render a \textit{position image} $J$ whose pixels store positional information about the scene points hit by the rays traced from the respective pixel. Since the pose of the calibration pattern and thereby the 3D positions of its points of interest are exactly known, the search for the ground truth positions within the rendering $I$ is reduced to simply finding the pixel positions with the correct positional information in the rendering $J$. Since a straight forward implementation of this idea is computationally expensive, we also propose a second method in which a ray in the scene space is calculated for every pixel, which is then intersected with the desired calibration pattern model to find the ground truth point positions. In summary, our contributions are:
\begin{itemize} \itemsep0em
	\item An extension of the plenoptic camera model of \cite{michels2018simulation} to include multiple microlens types
	\item Two methods for calculating the ground truth positions of the points of interest in the rendered images
	\item Publicly available implementations of the simulation and ground truth creation methods\footnote{\href{https://gitlab.com/ungetym/plenoptic_ground_truth_creator}{https://gitlab.com/ungetym/plenoptic\_ground\_truth\_creator}}
\end{itemize}
Note, that while the descriptions throughout this work are focused on plenoptic cameras, the whole pipeline is directly applicable to conventional cameras. We simply choose to describe the method for plenoptic cameras since these present a more complex case and the research in this area is in greater need of ground truth data as the standard approach of distorting perfect renderings is not applicable here.

\section{Related Work}
\textbf{Camera simulation in computer graphics:} The idea of using more realistic, physically-based lens models instead of a perfect pinhole camera for synthesizing images via ray tracing has first been explored by Potmesil and Chakravarty \cite{potmesil1981lens} and was later refined by Kolb \etal \cite{kolb1995realistic} and Wu \etal \cite{wu2010realistic}. These models were further extended by Wu \etal \cite{wu2013rendering} regarding the use of spectral ray tracing to simulate certain wave optics effects.\\
In contrast to these advanced methods for conventional cameras, the simulation of plenoptic cameras is a less explored area. This type of camera has only gained interest during the past decade due to the emergence of the first prototypes by Ng \etal \cite{ng2005lightfieldcamera} and the commercial realizations by Lytro (no longer existing) and Raytrix \cite{raytrix}. Despite becoming a more active field of research, the simulation of plenoptic cameras in most publications concerned with using synthetic images is rather rudimentary. Fleischmann \etal \cite{fleischmann2014plenoptic} render images without any main lens and Zhang \etal \cite{zhang2015forwardsimulation} as well as Liang \etal \cite{liang2015simuWOmainlens} use a simplified thin main lens model. Accordingly, the synthesized images do not show the distortion and image degradation effects present in real data. Further works by Liu \etal \cite{liu2015simulation} and Li \etal \cite{li2017numericalsimulation} based on ray splitting require an unrealistic large distance between the camera and the scene objects as well as simple scene materials. More recently, Michels \etal \cite{michels2018simulation} proposed to fully model a plenoptic camera's components and presented an implementation for Blender \cite{blender}. Despite wave optic effects and multiple microlens types not being simulated in this approach, we decided to base our method on it due to its availability and extensibility.\\
\textbf{Evaluation of pattern detectors:} While previous works on the calibration of plenoptic cameras use either manually labeled data \cite{nousias2017corner} or evaluate the detection and calibration as a combined system relying on precise real life measurements \cite{johannsen2013calibration}\cite{bok2017geometric}, approaches for conventional cameras have been evaluated with synthetic data for a broad variety of different patterns over the past decades. Luccese and Mitra \cite{lucchese2002using} use projective warping and Gaussian blur on checkerboard images and Zhang \cite{zhang2000flexible} renders square pattern images assuming a pinhole camera with non-zero skew and also applies Gaussian blur. Heikkila \cite{heikkila2000geometric} employs ray tracing and additional Gaussian noise as well as blur, but uses exactly the same camera model for rendering the point patterns as for the calibration. Ha \etal \cite{ha2017deltille} render sharp single triangle pattern corners for a perspective camera and add different levels of Gaussian noise and blur afterwards.\\
In summary, there is no previous work for the calibration of plenoptic cameras featuring synthetic data and the work dealing with conventional cameras uses simplified or ideal models to generate synthetic data.

\section{Organization}
Sections \ref{section_simulation} to \ref{section_mean_ray} describe the extended camera model for ray tracing and our general approaches for creating the ground truth positions. Subsequently, some insights regarding the usefulness of the direct approach via forward ray tracing are provided in section \ref{section_how_not_to}. Finally, in section \ref{section_realization} the realization for Blender is explained and the remaining sections are devoted to the evaluation of our approach.

\section{Method}\label{section_method}
The ray tracing approach presented in \cite{michels2018simulation} is capable of producing realistic images for plenoptic cameras with one microlens type and by deactivating the MLA it can also be used to simulate conventional cameras within the bounds of ray tracing, \ie without wave optical effects. Nevertheless, instead of directly using this simulation for our positional rendering approach, we first extend the camera model in order to also be able represent multifocus plenoptic cameras as distributed by Raytrix \cite{perwass2012single}.

\subsection{Simulation of Plenoptic Cameras}\label{section_simulation}
\begin{figure}[!t]
	\centering
	\includegraphics[width=.48\textwidth]{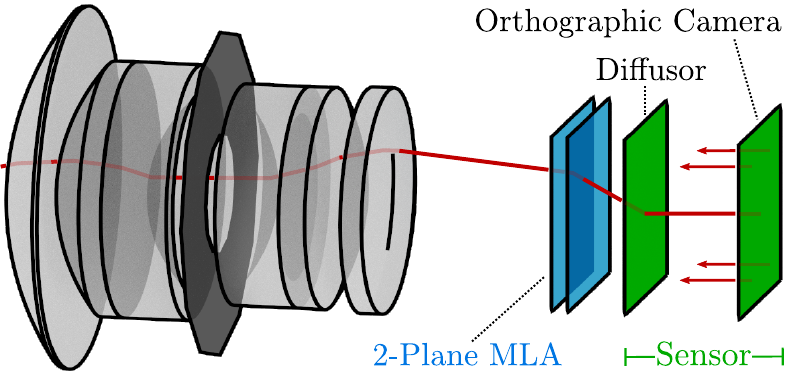}
	\caption{Schematics of the plenoptic camera model for ray tracing. While the objective's lenses are fully modeled, the MLA is approximated by two planes with recalculated normals and the sensor is simulated by a combination of an orthographic camera and a diffusor plane.}
	\label{fig:plenoptic_blender}
\end{figure}
\begin{figure}[!t]
	\centering
	\includegraphics[width=.48\textwidth]{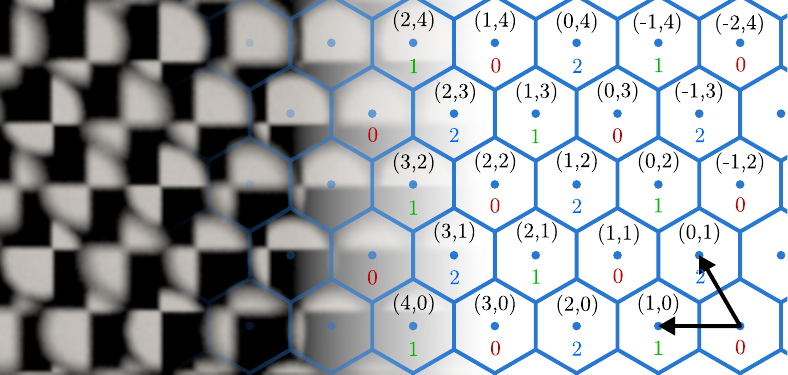}
	\caption{Rendering of a checkerboard with three differently focused microlens types overlaid with the hexagonal MLA layout. The black tuples are the coordinates of the center points with respect to the visualized base and the colored numbers indicate the lens type.}
	\label{fig:grid}
\end{figure}

The basic setup of the camera model is given in \autoref{fig:plenoptic_blender}. As described in \cite{michels2018simulation}, the objective's lenses are explicitly modeled and the refraction at their surfaces is smoothed by recalculating the surface normals in order to avoid image artifacts resulting from the polygonal surface structure. The sensor is modeled by combining an orthographic camera with a diffusor plane which randomly refracts the rays traced from the orthographic camera within a specified angle distribution. This simulates a real pixel's field of view (FOV) and its response to light rays with different angles of incidence. Accordingly, the diffusor plane can be thought of as the location of the sensor.\\
The last component, the MLA, is designed as a simple two plane model by exploiting the lensmaker's equation for a thin lens with index of refraction (IOR) $\eta$ and focal length $f$, given by
\begin{equation}
\frac{1}{f}\approx (\eta-1)\left(\frac{1}{R_1}-\frac{1}{R_2}\right),
\end{equation}
where $R_1$ and $R_2$ describe the front and back surface curvature radii. For a flat back surface, given by $R_2=\infty$, it follows $f\approx R_1/(\eta-1)$ and since a large radius $R_1$ leads to the front surface locally nearly being a plane, the microlenses can be constructed by using a two plane model with large IOR $\eta$ and recalculated surface normals. We extend this part of the model to feature differently focused microlenses on the same MLA as shown in \autoref{fig:grid} by setting different values for $R_1$ depending on the coordinates of a microlens in the hexagonal grid. 
Since to our knowledge the Raytrix cameras are the only commercially available multi focus plenoptic cameras, we describe the extension for a setup with three microlens types as used by Raytrix. This model, however, can easily be modified to feature different configurations.\\
For the three microlens setup the type $t\in\{0,1,2\}$ of a microlens with center coordinates $(i,j)$ in the hexagonal grid is given by $t=(i-j)\%3$ as visualized in \autoref{fig:grid}. In order to match the setup of a Raytrix camera, the three focal lengths $R_1^t$ have to be chosen carefully with two restrictions in mind. First, all focal lengths should be larger than the distance between the MLA and the sensor plane since the MLA in Raytrix cameras is placed between the main lens and the virtual image of the scene, thus the microlenses collect converging light rays (compare \autoref{fig:plenoptic}). And second, the depth of field (DoF) of the three lens types should slightly overlap to create a connected combined DoF without focus gaps \cite{perwass2012single}.\\
\noindent
The described model can now be used to render realistic images of calibration patterns (or arbitrary scenes) for various plenoptic as well as conventional camera setups, where the camera type can be switched by (de)activating the MLA and modifying the parameters and positioning of the sensor and MLA.
Note, that for the sake of simplicity, the illustrations in the remaining sections will contain the schematics of a real plenoptic camera instead of the model described here.

\subsection{Rendering Positional Information}\label{section_positional}
\begin{figure*}[t]
	\centering
	\includegraphics[width=0.99\textwidth]{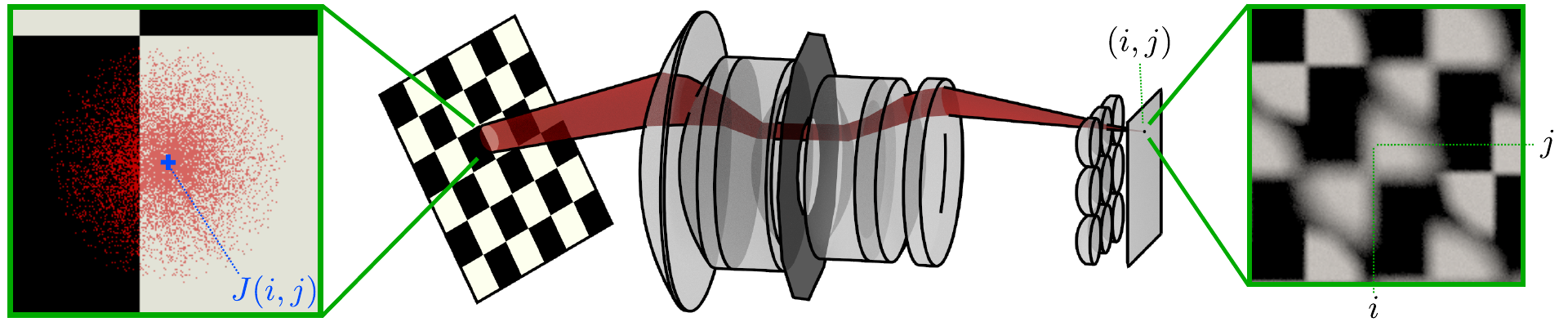}
	\caption{Positional rendering visualized: The right image shows a section of the rendering $I$ containing the pixel $(i,j)$ and the schematics in the middle visualize the bundle of rays traced from $(i,j)$ and its intersection with the calibration pattern object. On the left the set of scene points hit by the rays, $\{p(r):r\in R^{(i,j)}_{hit}\}$, is shown in red. Despite the calibration pattern not being in focus of the microlens, the pixel's positional information, $J(i,j)$, can be calculated as the mean of the set of points, shown in blue.}
	\label{fig:positional}
\end{figure*}
\begin{figure}[!t]
	\centering
	\includegraphics[width=0.9\linewidth]{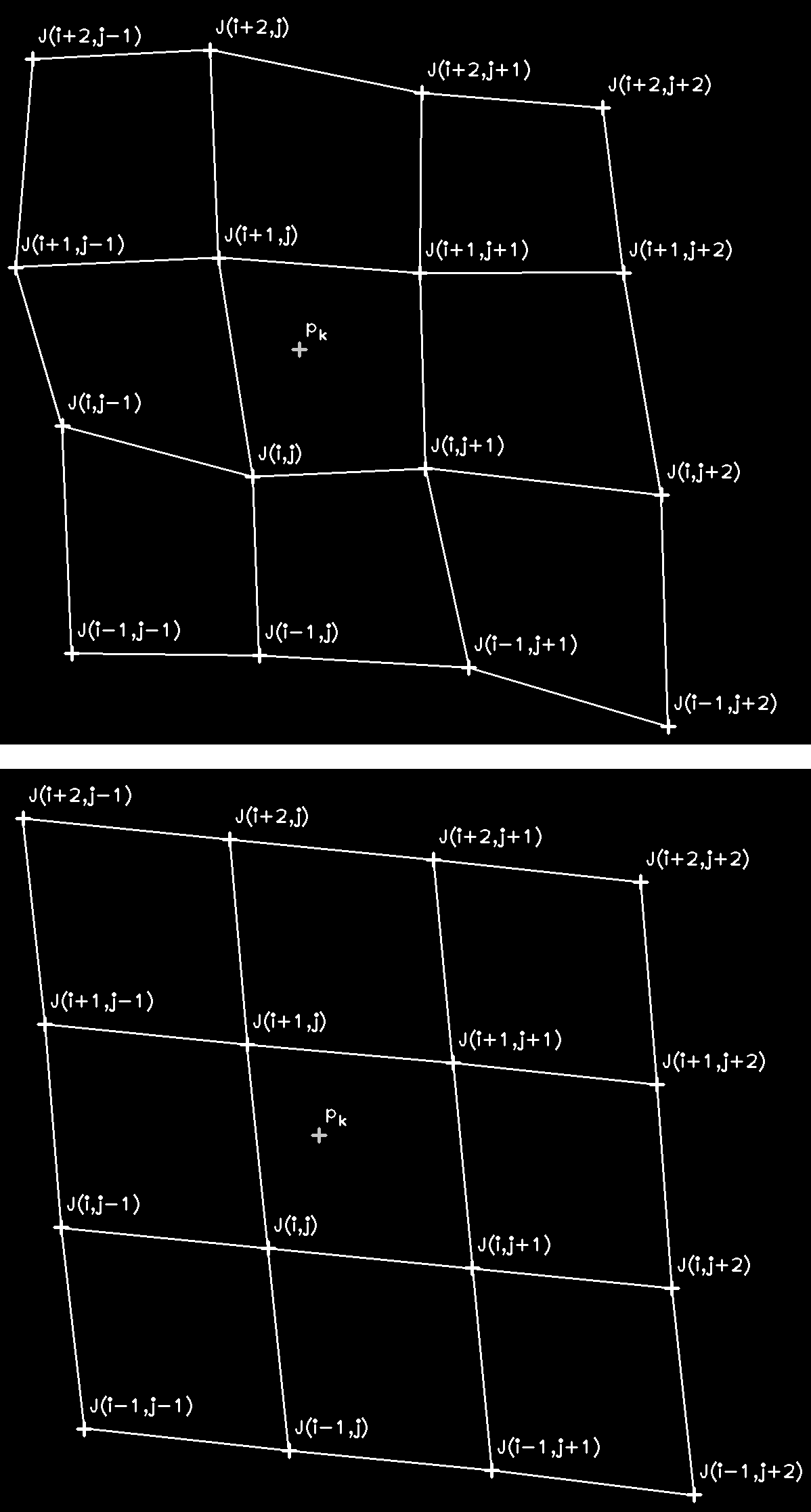}
	\caption{The values of $J$ in a neighborhood of $(i,j)$ approximately form a grid on the calibration pattern plane. This visualization shows the effect that the number of rays has on the grid structure. The positional images $J$ used here were calculated with $64^2$ (top) and $256^2$ (bottom) samples per pixel.}
	\label{fig:grids_compare}
\end{figure}

Since the pattern position and orientation are exactly known for the rendering, we can assume to have a set $\{p_k\}_{k=1,\cdots,n}\subset\mathbb{R}^3$ of locations of the $n$ relevant pattern points, \eg the $n=28$ corners of a $4\times7$ checkerboard. In order to use this information for finding the ground truth pixel positions of the calibration pattern points in the images rendered with the previously described setup, the same model is used to render positional information via ray tracing. In the usual backward ray tracing pipeline, for every pixel $(i,j)$ a set of rays $R^{(i,j)}$ is traced through the camera into the scene and the colors of the scene points hit by the rays are accumulated which will be shortly formalized in the following. Let $p(r)\in \mathbb{R}^3$ denote the first scene point hit by the ray $r\in R^{(i,j)}$ and split the set of rays into two disjoint sets $R^{(i,j)}=R^{(i,j)}_{blocked}\ \dot{\cup}\ R^{(i,j)}_{hit}$ with $R^{(i,j)}_{blocked}$ containing the rays not leaving the camera due to being blocked by the aperture or camera housing and $R^{(i,j)}_{hit}$ denoting the set of rays intersecting scene objects, whereby every ray leaving the camera is assumed to be in $R^{(i,j)}_{hit}$. The color of a pixel $(i,j)$ in the calibration pattern rendering $I:\{0,\cdots,width\}\times\{0,\cdots,height\}\rightarrow\{0,\cdots,255\}^3$ is then given by
\begin{align}\label{eq:I}
I(i,j) &= \frac{1}{|R^{(i,j)}|}\sum_{r\in R^{(i,j)}}c(p(r))\nonumber\\
&= \frac{1}{|R^{(i,j)}|}\sum_{r\in R^{(i,j)}_{hit}}c(p(r)),
\end{align}
where $c(p(r))$ describes the color of the 3D point hit by ray r and rays in $R^{(i,j)}_{blocked}$ are not assumed to add non-zero color information. We would like to remark, that the color $c(p(r))$ can be the result of further ray tracing calculations depending on the scene objects' reflectivity and transmission properties. However, for the task at hand only the first scene point hit by a ray, $p(r)$, is considered.\\
For the positional rendering the same procedure is used, but instead of the color values $c(p(r))$ the positions $p(r)$ are accumulated and averaged, \ie the value of the positional rendering $J:\{0,\cdots,width\}\times\{0,\cdots,height\}\rightarrow\mathbb{R}^3$ at pixel position $(i,j)$ is given by 
\begin{equation}\label{def_J}
J(i,j) = \frac{1}{\left|R_{hit}^{(i,j)}\right|}\sum_{r\in R^{(i,j)}_{hit}}p(r),
\end{equation}
as visualized in \autoref{fig:positional}. Note, that the pixel value is normalized by $\left|R_{hit}^{(i,j)}\right|$ instead of $|R^{(i,j)}|$ as in \autoref{eq:I} since we are interested in the average scene point hit by the rays unbiased by vignetting, \ie the amount of blocked rays.\\
Despite knowing the average scene position $J(i,j)$ a camera pixel $(i,j)$ is seeing, the known 3D calibration point positions $\{p_k\}$ can most likely not directly be found in the positional rendering due to $J$ maximally containing $width\times height$ 3D positions of the continuous calibration pattern plane. The naive solution to this problem is searching for pixels at which the value of $J$ is close to a position $\{p_k\}$, \ie for every $p_k$ we search for
\begin{equation}\label{eq:naive_solution}
(\hat{\imath},\hat{\jmath})=\arg\min_{(i,j)}||p_k-J(i,j)||
\end{equation}
and accept the solution $(\hat{\imath},\hat{\jmath})$, if the distance for this position is within a certain threshold, $||p_k-J(\hat{\imath},\hat{\jmath})||<\lambda$ for some $\lambda>0$.
This procedure, however, does not work for plenoptic images since these can contain a scene point multiple times in different microlens images as shown in \autoref{fig:plenoptic} and \autoref{fig:grid}. Fortunately, the MLA configuration is known and therefore the image $J$ can be splitted into microlens images $J_1,\cdots,J_m$, each containing only the rendered information for exactly one microlens. In these images the search can then independently be performed.\\
This naive solution for finding the ground truth pixel positions has the obvious drawback of a limited accuracy. The solution is only accurate within $\pm 0.5 px$ and if the number of samples, \ie rays per pixel, is not sufficient, the found pixel $(\hat{\imath},\hat{\jmath})$ might even be an outlier due to $J(\hat{\imath},\hat{\jmath})$ containing a wrong position. This accuracy problem will be tackled in the following by rendering $J$ with a higher resolution than $I$, filtering out unreliable results and finally using interpolation near the filtered pixel positions. First, one can observe, that for a sufficiently large number of samples per pixel, the values of small neighborhoods in $J$ form an equidistant grid on the calibration pattern plane as visualized in \autoref{fig:grids_compare}. 
This observation is used as a constraint for filtering the point candidates.\\
Assume $J$ was rendered with a resolution of $K\cdot width\times K\cdot height$, $K\in\mathbb{N}$ and let $(\hat{\imath},\hat{\jmath})$ a pixel such that some corner position $p_k$ is located in the polygon given by $J(\hat{\imath},\hat{\jmath}),J(\hat{\imath}+1,\hat{\jmath}),J(\hat{\imath},\hat{\jmath}+1)$ and $J(\hat{\imath}+1,\hat{\jmath}+1)$ as shown in \autoref{fig:grids_compare} and without loss of generality let $J(\hat{\imath},\hat{\jmath})$ be the closest of the four corners to $p_k$. In order to allow the interpolation of the pixel position within these coordinates, we first check, if the neighborhood $N:=\{(\hat{\imath}+a,\hat{\jmath}+b):-1\leq a\leq 2, -1\leq b \leq 2\}$ approximately forms an equidistant grid. To this end, the average distances between the values of vertical and horizontal neighbors,
\begin{align}
d_{vert} &= \frac{1}{12}\sum_{(i,j)\in N|_{b<2}}||J(i,j)-J(i,j+1)||\ \text{and}\\
d_{horiz} &= \frac{1}{12}\sum_{(i,j)\in N|_{a<2}}||J(i,j)-J(i+1,j)||,
\end{align}
are calculated and then used to define the first constraint
\begin{equation}
\left|1-\frac{||J(i,j)-J(i,j+1)||}{d_{vert}}\right|<\lambda_d
\end{equation}
for all $(i,j)\in N|_{b<2}$ and a threshold $\lambda_d\in(0,1)$ and analogously
\begin{equation}
\left|1-\frac{||J(i,j)-J(i+1,j)||}{d_{horiz}}\right|<\lambda_d
\end{equation}
for all $(i,j)\in N|_{a<2}$. In this length constraint $\lambda_d$ describes the maximal relative deviation which simply enforces that the lengths of horizontal and vertical lines in the grid do not deviate too much from the respective average. A similar constraint is calculated for the angles of grid connections, \ie the average angle 
\begin{equation}
\alpha = \frac{1}{9}\sum_{(i,j)\in N|_{a<2,b<2}}\alpha_{(i,j)}
\end{equation}
with $\alpha_{(i,j)}:=\angle(J(i,j),J(i+1,j),J(i,j+1))$ is calculated and the respective constraint is formulated as
\begin{equation}
\forall (i,j)\in N|_{a<2,b<2}: |\alpha_{(i,j)} - \alpha|<\lambda_\alpha
\end{equation}
for a threshold $\lambda_\alpha\in(0,\pi)$. If both constraints hold for the neighborhood of $(\hat{\imath},\hat{\jmath})$, the ground truth pixel position $(\tilde{\imath},\tilde{\jmath})$ in $I$ is interpolated via
\begin{equation}
(\tilde{\imath},\tilde{\jmath}) = \frac{1}{K}\left((\hat{\imath},\hat{\jmath})+s\cdot (1,0)+t\cdot (0,1)\right)
\end{equation}
where $s$ and $t$ are the solution of the linear equation
\begin{align}
p_k=J(\hat{\imath},\hat{\jmath})&+s(J(\hat{\imath}+1,\hat{\jmath})-J(\hat{\imath},\hat{\jmath}))\nonumber\\
&+t(J(\hat{\imath},\hat{\jmath}+1)-J(\hat{\imath},\hat{\jmath}))
\end{align}
and $\frac{1}{K}$ is used to rescale the pixel position to the size of $I$. Note, that the solution $(s,t)$ exists and does not require numerical approximations since all values of $J$ as well as the point $p_k$ are located on the same plane.

\subsection{Calculating Pixel Rays}\label{section_mean_ray}
\begin{figure}[!t]
	\centering
	\includegraphics[width=.48\textwidth]{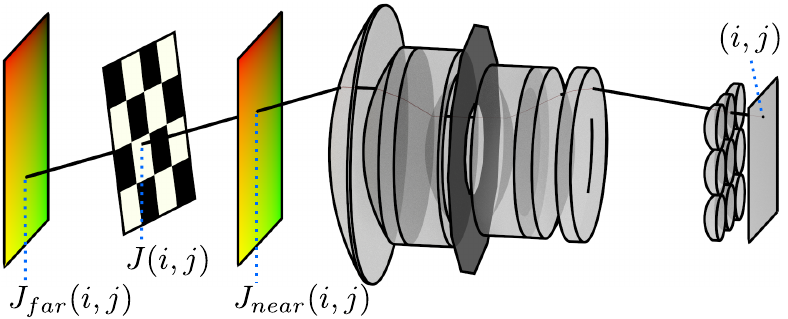}
	\caption{Two plane approach: The colored planes are used to create positional renderings $J_{near}$ and $J_{far}$ and the positional information $J(i,j)$ for a pixel $(i,j)$ is then given by the intersection of the calibration pattern object and the ray defined by $J_{near}(i,j)$ and $J_{far}(i,j)$.}
	\label{fig:twoplane}
\end{figure}

A major disadvantage of the approach described in the previous section is, that it requires one additional image to be rendered for every calibration pattern position. Especially for a large resolution scaling factor $K$ and large sample numbers this is inefficient considering that the camera setup usually does not change during the creation of one dataset and therefore the exact same rays are traced through the camera into the scene for every positional rendering. To circumvent this redundancy, we propose the rendering of only two positional images per camera setup - the first one, $J_{near}$, for a plane located at the start of the cameras DoF and another one, $J_{far}$, for a plane at the DoF's end. For every pixel $(i,j)$ the 3D points $J_{near}(i,j)$ and $J_{far}(i,j)$ define a ray in the scene space (compare \autoref{fig:twoplane}) similar to the often used two-plane parametrization of the plenoptic function. Given the rendering of a calibration pattern $I$ as before, the corresponding positional image $J$, as defined in the previous chapter, can be calculated by intersecting the pattern plane and the pixel rays, \ie
\begin{align}
J(i,j) &= J_{near}(i,j)+t\cdot(J_{far}(i,j)-J_{near}(i,j))\\[8pt]
with&\quad t =\frac{<q-J_{near}(i,j),n>}{<J_{far}(i,j)-J_{near}(i,j),n>}
\end{align}
where $q$ is an arbitrary point of the calibration pattern plane and $n$ denotes its normal.\\
This method for calculating the positional image $J$ requires only two positional renderings per camera setup instead of one rendering per calibration pattern image $I$.

\subsection{How-Not-To: Forward Ray Tracing}\label{section_how_not_to}
Instead of rendering whole positional images using computationally expensive backward ray tracing with large numbers of samples, one might wonder why we do not simply use forward ray tracing, \ie tracing rays from the known positions $\{p_k\}$ to the sensor. This idea is appealing since the number of required rays would be heavily reduced. However, this approach only works for scene points that are either in focus or create a perfect circle of confusion on the sensor. In the former case, the rays all hit exactly one single pixel on the sensor (for a plenoptic camera with one microlens type, they might hit unique pixels in different microlens images) and for the latter one can simply take the center of the circle of confusion as the ground truth position. By treating the sensor as a continuous plane instead of discretizing it into pixels during the ray tracing, even sub-pixel accuracy could be reached. However, the shape of the area of confusion can vary heavily depending on the optical system used for the imaging and it is unclear, which point could be regarded as ground truth position for arbitrary shapes which in addition can be split over multiple microlens images.\\
Nevertheless, the forward ray tracing could be used to determine the areas of the sensor which should be rendered via backward ray tracing. After rendering a calibration pattern image $I$, the positions $\{p_k\}$ could be traced to the continuous sensor plane and after choosing the resolution of $J$, these sensor areas could be discretized into a set of pixels which are subsequently used for the positional rendering. 

\section{Evaluation}\label{section_evaluation}
\begin{figure*}[!t]
	\centering
	\includegraphics[width=0.99\textwidth]{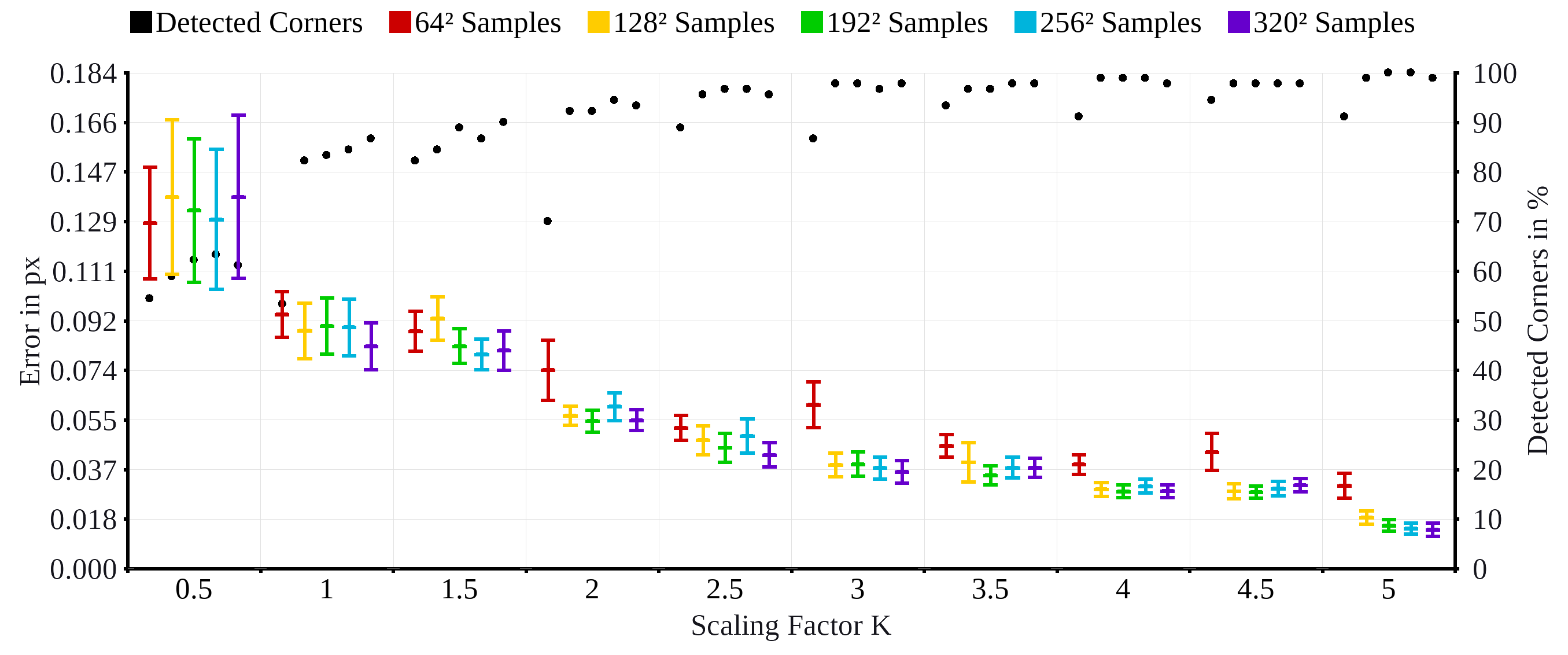}
	\caption{Method 1: Accuracy of interpolated corners for different numbers of samples and different scalings of $J$. For every combination of a scaling factor and a number of samples the resulting corners were compared to the reference solution rendered with $K=10$ and $102400$ samples. The average difference and standard error of mean (SEM) in terms of pixels are shown by the colored bars. Furthermore, the black dots show the ratio of detected corners corresponding to the respective colored bars.}
	\label{fig:eval_2d}
\end{figure*}

\subsection{Realization in Blender}\label{section_realization}
In order to evaluate our approach, first the model of \cite{michels2018simulation} for Blender 2.79c was extended by modifying the MLA materials to support up to three configurable microlens types as described in section \ref{section_simulation}. With this setup, a calibration pattern rendering $I$ can easily be created. A corresponding positional image $J$, however can not directly be rendered since the Cycles renderer does not provide the functionality to only accumulate rays hitting the scene. However, giving the calibration pattern plane a material that emits positional information, \ie $c(p(r))=p(r)$, and everything else a purely black material results in a rendering $\hat{J}$ with
\begin{equation}
\hat{J}(i,j) = \frac{1}{\left|R^{(i,j)}\right|}\bigg(\sum_{r\in R^{(i,j)}_{hit}}p(r)+\underbrace{\sum_{r\in R^{(i,j)}_{blocked}}p(r)}_{=0}\bigg)
\end{equation}
which differs from $J$ (see \autoref{def_J}) only by the factor $|R^{(i,j)_{hit}}|/|R^{(i,j)}|$ describing the ratio of rays hitting the calibration pattern. This factor can be calculated by rendering an additional single channel image $J_{white}$ for which the calibration pattern plane emits a purely white material and everything else remains black, \ie $c(p(r)) = 1.0$ if $r\in R^{(i,j)}_{hit}$ and $c(p(r)) = 0.0$ otherwise. The resulting image $J_{white}$ then contains the desired factor,
\begin{equation}
J_{white}(i,j) = \frac{1}{\left|R^{(i,j)}\right|}\bigg(\underbrace{\sum_{r\in R^{(i,j)}_{hit}}1}_{=|R^{(i,j)_{hit}}|}+\underbrace{\sum_{r\in R^{(i,j)}_{blocked}}0}_{=0}\bigg)
\end{equation}
and accordingly $J(i,j)$ can be calculated by dividing the three channels of $\hat{J}(i,j)$ by $J_{white}(i,j)$.\\ Unfortunately, this procedure requires a lot of redundant ray tracing since the same rays are traced into the scene for $\hat{J}$ as for $J_{white}$. Fortunately, a set point $p\in\mathbb{R}^3$ on a plane parameterized by $a+u\cdot b+v\cdot c$ with $a,b,c\in\mathbb{R}^3$ is uniquely determined by the parameters $(u_p,v_p)$ with $a+u_p\cdot b+v_p\cdot c=p$. Thus the rendering of UV coordinates suffices to reconstruct the corresponding 3D point on the plane. Consequently only two channels are needed to save the positional information and the third channel of $\hat{J}$ can be used to store the ray proportion $J_{white}$. Analogous to the previous normalization, the positional image in terms of UV coordinates, $J_{UV}$ is given by dividing the first two channels of $\hat{J}$ by its third channel. The search for ground truth positions can then be performed by transferring $\{p_k\}$ into UV coordinates and searching these in $J_{UV}$.\\
For the second method the reparametrization of the planes is not necessary. Placing the two planes parallel to the worlds coordinate axes results in the points of the same plane having one fixed coordinate. This fixed coordinate can be saved in a small configuration file instead of the image channels of $J_{near}$ and $J_{far}$, thus the freed channel in these images can again be used to save the ray proportion $J_{white}$. The final positional rendering $J_{near}$ and $J_{far}$ are then calculated by dividing the two positional channels by the ray proportion channel and subsequently replacing the latter by the externally saved fixed coordinate.

\subsection{Number of Samples and Resolution}
In order to assess the dependency of the resulting accuracy on the number of samples and the positional image resolution, we used a plenoptic camera setup with a double Gaussian 100mm objective, an MLA-to-sensor distance of 1.7mm, an MLA-to-main lens distance of 123.3mm and focal lengths of 1.9mm, 2.1mm and 2.3mm for the microlenses. The MLA as well as sensor have a size of $21.73mm\times21.73mm$ and the MLA contains approximately $100\times115$ microlenses whereby the larger number of microlenses in the vertical is a result of the hexagonal ordering of the lenses. Furthermore, the thresholds for the constraints given in section \ref{section_positional} have been empirically chosen as $\lambda_d=0.15$ and $\lambda_\alpha=10^\circ$. We would like to remark, that further tests confirmed, that the general conclusions of the following evaluation also hold for different threshold values as these mainly regulate the number of positional outliers.\\
With this setup we rendered multiple images showing a checkerboard located at different depths with varying angles. For these images we applied our first approach for different combinations of sample numbers and resolutions. The results are presented in \autoref{fig:eval_2d} where the mean and SEM of the differences between the determined pixel positions and the reference positions is shown. These results show, that increasing the scaling factor $K$ significantly decreases the error and variance while the number of detected corners significantly improves. In contrast to this observation, the number of samples seems to have only a limited impact in the tested range. The results rendered with $320^2$ samples show an average improvement of $0.0028\ px$ for the pixel positions and $1.6\%$ for the ratio of detected corners, compared to the results produced with $128^2$ samples. Only positional images $J$ rendered with significantly fewer samples seem to suffer from an imprecision caused by the low sample rate as the results produced with $64^2$ samples suggest.

\subsection{Comparison of the two Methods}
\begin{figure*}[!t]
	\centering
	\includegraphics[width=0.95\textwidth]{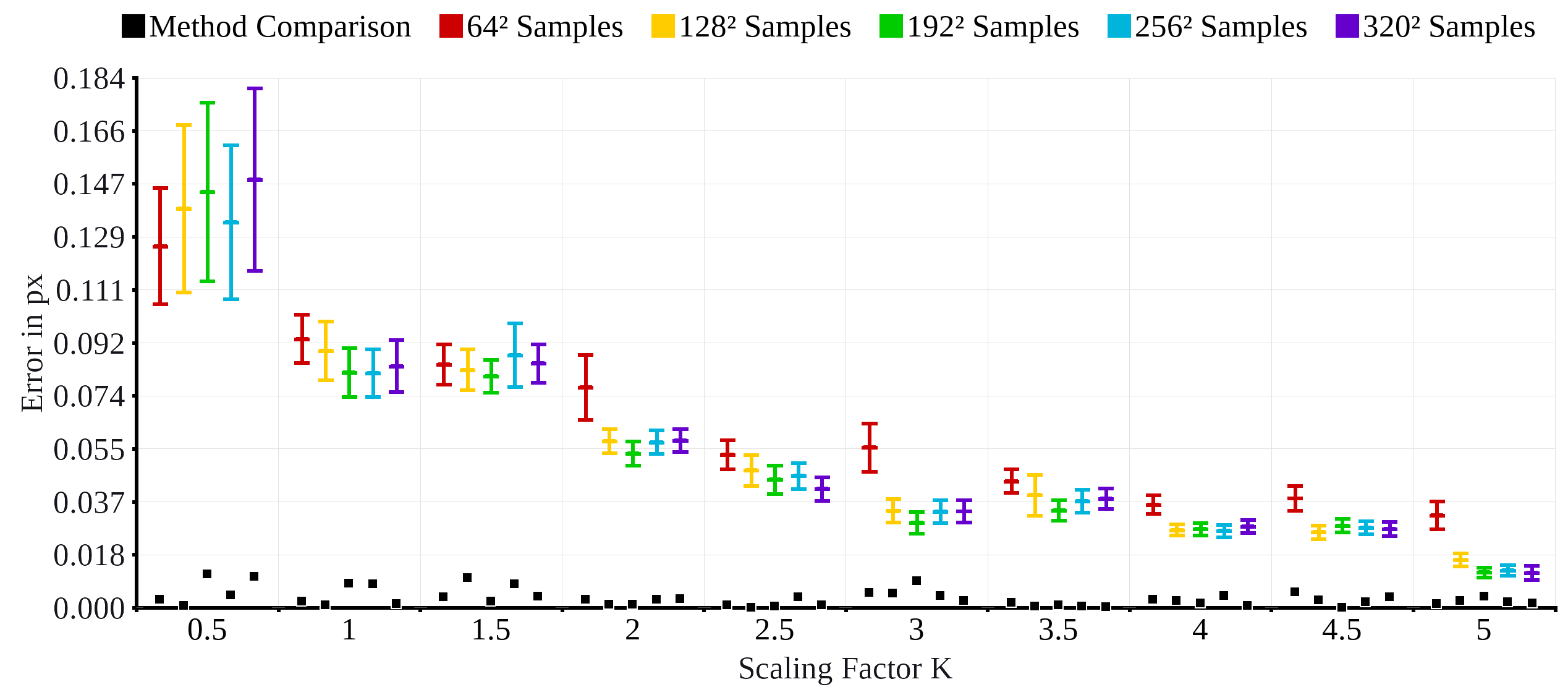}
	\caption{Method 2: Accuracy of corners calculated via the two plane method. The resulting corners were compared to the same reference solution as in \autoref{fig:eval_2d}. The average difference and SEM in terms of pixels are shown by the colored bars and the black squares show the absolute difference between the means of both methods for the respective combination of $K$ and number of samples.}
	\label{fig:eval_2d_comp}
\end{figure*}
The same combinations of sample numbers and resolutions shown in \autoref{fig:eval_2d} were also used to evaluate the differences between the two proposed approaches. As \autoref{fig:eval_2d_comp} shows, the convergence behavior of the two plane approach with respect to the reference from the previous section is identical to that of the first method and the means of both methods only deviate from each other by less than less than $0.016\ px$. The remaining fluctuations between the two methods, which show no clear winner in regards of accuracy, are a result of two aspects. Firstly, the two plane method has the disadvantage of additional intersection calculations which can introduce further precision errors, especially since the ray tracing is usually done on GPUs with only single precision. And second, small positional errors can have different effects in both methods. While this error is located directly on the calibration pattern plane for the first method, the impact of a positional error in the two plane method varies with the pose of the calibration pattern since the error is located on the near or far plane.\\
However, since the difference in accuracy is negligible for a sufficiently large number of samples, the two plane method is recommended due to the significantly smaller render time. In our experiments we used two Nvidia Titan X and with that setup, the rendering of an image $I$ or $J$ with a resolution of $width\times height$ pixel and $n$ samples per pixel took approximately $t \approx width\cdot height \cdot n \cdot 10^{-11}$ minutes. Furthermore, the time needed for searching the positions $\{p_k\}$, including the calculation of $J$ from $J_{near}$ and $J_{far}$ in the second method, is in both cases by several orders of magnitude smaller than the rendering time. Accordingly, the two plane method is significantly faster for every dataset consisting of more than two images per camera setup.

\subsection{Conclusion and Limitations}
The proposed methods are able to produce highly accurate, realistic data for the evaluation of calibration methods and a wide range of cameras. However, they are limited regarding the geometry of the calibration object. Throughout this work, it is assumed that the calibration pattern is placed on a plane, which excludes calibration objects, like checkerboard cubes, featuring a three dimensional structure. If the scene points $\{p(r)\}$ that are hit by the rays $r\in R^{(i,j)}$ traced from the pixel $(i,j)$ are not located on a common plane, the calculation of $J(i,j)$ via simple averaging as described in \autoref{def_J} is incorrect. To a limited extent this problem might be avoidable by using the two plane method. However, it remains to be analyzed whether the intersection of a more complex scene and the mean ray of a pixel can be used in the same manner as in this work.\\
Another problem of geometrical nature results from the proposed method for filtering outliers, where it is assumed, that a small neighborhood of pixels is projected to a grid on the calibration pattern plane as shown in \autoref{fig:grids_compare}. While this assumption holds true for most common camera setups, it is theoretically possible for an optical system to contain high frequency distortions which deform the grids even in the smallest neighborhoods. In this case it is recommended to skip the filter and simply render $J$ or the two proxy planes with a significantly larger resolution such that the solution of the naive approach given by \autoref{eq:naive_solution} is sufficiently accurate.\\

\section{Acknowledgment}
This work was supported by the German Research Foundation, DFG, No. K02044/8-1 and the EU Horizon 2020 program under the Marie Sklodowska-Curie grant agreement No 676401.

{\small
\bibliographystyle{IEEEtran}
\bibliography{bibliography/lit}
}

\end{document}